\pdfoutput=1
\documentclass[11pt]{article}

\usepackage{EMNLP2023}

\usepackage{times}
\usepackage{latexsym}

\usepackage[T1]{fontenc}
\usepackage[utf8]{inputenc}

\usepackage{microtype}

\usepackage{inconsolata}
\usepackage{amsmath} 
\usepackage{graphicx}
\usepackage{booktabs}

\title{Beyond WER: Probing Whisper's Sub‑token Decoder Across Diverse Language Resource Levels}

\author{
  Siyu Liang$^{1}$,
  Nicolas Ballier$^{2}$,
  Gina-Anne Levow$^{1}$,
  Richard Wright$^{1}$ \\
  $^{1}$ Department of Linguistics, University of Washington \\
  $^{2}$ ALTAE, Université Paris Cité,  F-75013 Paris, France \\
  \texttt{\{liangsy, levow, rawright\}@uw.edu, nicolas.ballier@u-paris.fr}
}

\begin{document}
\maketitle
\begin{abstract}
While large multilingual automatic speech recognition (ASR) models achieve remarkable performance, the internal mechanisms of the end-to-end pipeline, particularly concerning fairness and efficacy across languages, remain underexplored. This paper introduces a fine-grained analysis of Whisper’s multilingual decoder, examining its sub-token hypotheses during transcription across languages with various resource levels. Our method traces the beam search path, capturing sub-token guesses and their associated probabilities. Results reveal that higher resource languages benefit from higher likelihood of the correct token being top-ranked, greater confidence, lower predictive entropy, and more diverse alternative candidates. Lower resource languages fare worse on these metrics, but also exhibit distinct clustering patterns in sub-token usage sometimes influenced by typology in our PCA and t-SNE analysis. This sub-token probing uncovers systematic decoding disparities masked by aggregate error rates and points towards targeted interventions to ameliorate the imbalanced development of speech technology.
\end{abstract}

\section{Introduction}
Large multilingual automatic speech recognition (ASR) models like Whisper \citep{radford_robust_2023} demonstrate impressive capabilities on high-resource languages, yet their performance often degrades significantly for low-resource languages \citep{javed_towards_2022}, alongside persistent concerns about fairness across diverse linguistic groups \citep{zee_group_2024}. Aggregate metrics such as Word Error Rate (WER) can obscure the nuanced ways these models falter internally and may not capture critical issues like hallucination \citep{koenecke_careless_2024}. This necessitates deeper analysis of the decoding process itself, with some prior work also highlighting the value of evaluating models at the sub-unit level, for instance, in assessing calibration \citep{ballier_probing_2024}.

This paper posits that a granular, sub-token level investigation of Whisper's decoder is crucial for a more comprehensive understanding of these performance variations. We use the term `sub-token' to refer to the sub-word units (e.g., Byte Pair Encoding (BPE) units) that models like Whisper generate, often broadly called `tokens' in the literature. Recognizing that tokenization strategies can themselves introduce biases and affect model behavior \citep{petrov_language_2023, ahia_magnet_2024}, we scrutinize how key characteristics of the sub-token generation process systematically differ when processing languages with varying training resources: specifically, the rank of chosen sub-tokens, model confidence in its predictions, predictive uncertainty (entropy), the diversity of the hypothesis space, and overall sub-token usage patterns.

Our analysis empirically demonstrates two primary findings: first, higher resource languages consistently benefit from more robust decoding metrics at the sub-token level, including higher prediction confidence and lower predictive entropy. Second, sub-token usage patterns as revealed through principal component analysis (PCA) and t-distributed stochastic neighbor embedding (t-SNE) indicate poorer handling of tokenization for lower resource languages, but also reveal typologically coherent clusters that can transcend simple resource-level distinctions, highlighting the interplay between linguistic structure and data availability. These fine-grained insights are valuable for developing targeted interventions, such as specialized adapter fine-tuning \citep{song_lora-whisper_2024, pfeiffer_unks_2021}, to improve the equity and efficacy of multilingual ASR systems.

\section{Background}
\subsection{Whisper and Tokenisation}
Whisper is an influential foundation encoder-decoder Transformer model \citep{radford_robust_2023}. It was trained using large-scale weak supervision on approximately 680,000 hours of multilingual audio data, covering a wide array of tasks, including speech transcription and translation. This extensive pre-training enables strong zero-shot performance.

A core component of Whisper's architecture, as well as many modern large language and speech models, is its tokenization strategy. As detailed by \citet{radford_robust_2023}, Whisper utilizes two separate Byte Pair Encoding (BPE) vocabularies: one derived from the GPT-2 tokenizer \citep{sennrich_neural_2016, radford_language_2019} for English-only models, and a distinct, refitted vocabulary of the same size for multilingual models. This refitting was intended to avoid excessive fragmentation on other languages since the BPE vocabulary is English only \citep{radford_robust_2023}. Our work focuses on the behavior of models using this multilingual BPE vocabulary, which is shared across all non-English languages the model supports. During decoding, the model generates a sequence of these sub-tokens, typically guided by special tokens like a language ID. While sub-word units allow handling large vocabularies and morphological variations more effectively than word-level tokenization, and can facilitate cross-lingual transfer \citep{conneau_unsupervised_2020}, \citet{radford_robust_2023} themselves acknowledge potential limitations, particularly for languages distant from the Indo-European family which forms the bulk of the training data. They note that performance outliers could be due to a lack of transfer across languages and that the BPE tokenizer could be a poor match for these languages or variations in data quality.

However, the nature of sub-word tokenization, especially in a multilingual context, is not without its challenges. The way texts are segmented into tokens can vary significantly across languages, potentially leading to disparities in processing efficiency, context window utilization, and even model performance \citep{petrov_language_2023}. For instance, some languages might be systematically broken into more tokens than others for equivalent semantic content, an issue explored in the context of text-based LLMs \citep{petrov_language_2023, acs_exploring_2019}. Such tokenization artifacts can contribute to unfairness, as models might inherently find it more complex to process or learn representations for languages that result in longer token sequences \citep{ahia_magnet_2024}. While recent research also explores discrete acoustic or semantic tokens for ASR \citep{guo_recent_2025, cui_exploring_2024}, the BPE approach as employed in Whisper remains a common paradigm, making the study of its sub-token characteristics critical.

\subsection{Beam Search Decoding}
For generating transcriptions, Whisper typically employs beam search decoding. Beam search maintains a set of $k$ (the beam width) most probable partial hypotheses (sequences of tokens). At each step, it extends these hypotheses with possible next tokens and re-ranks them based on their cumulative probabilities, pruning the set back to $k$ hypotheses. This exploration of multiple paths aims to find an overall sequence with a higher likelihood than what might be found through a purely greedy approach. Our analysis focuses on the characteristics observed along the single, final beam path chosen by the model, representing its ultimate transcription output. Probing model predictions, particularly the probability scores assigned by the decoder to chosen sub-tokens, provides insights into the model's decision-making at each step of this generation process. These output probabilities are commonly used as a direct measure of model confidence in the ASR literature \citep[e.g.,][]{jiang_confidence_2005, ballier_probing_2024, ballier_whisper_2024, aggarwal_adopting_2025}. However, it is also well-established that the raw softmax probabilities from deep neural networks may not always be well-calibrated and can exhibit overconfidence \citep{guo_calibration_2017}.

\subsection{Resource Disparity and Fairness in Multilingual ASR}
A significant challenge in developing truly equitable multilingual ASR systems is the vast disparity in available training resources across languages. While Whisper's pre-training dataset is exceptionally large, the distribution of data per language varies by orders of magnitude \citep{radford_robust_2023}. This imbalance directly impacts model performance, generally leading to superior results for higher resource languages \citep{javed_towards_2022}. Scaling models and data (e.g., \citet{tjandra_massively_2023} and \citet{pratap_scaling_2024}) can improve overall performance but does not necessarily resolve fairness issues or guarantee equitable performance across all languages and speaker groups \citep{zee_group_2024}. In fact, \citet{zee_group_2024} found that larger models can sometimes exhibit greater worst-case performance disparities. 

The challenges for low resource languages are multifaceted. Beyond raw data quantity, issues include the representation of diverse scripts \citep{pfeiffer_unks_2021, muller_when_2021}, the quality of sub-token vocabularies for these languages \citep{downey_embedding_2023}, and the potential for "capacity dilution" where a fixed-size model struggles to adequately represent many languages \citep{conneau_unsupervised_2020}. These factors can lead to higher error rates, lower model confidence, and increased susceptibility to issues like hallucination \citep{koenecke_careless_2024} for low resource languages. Prior work often evaluates these disparities at the word or utterance level (e.g., WER), with dedicated studies benchmarking performance on specific low-resource language sets like Pashto, Punjabi, and Urdu \citep{sehar_benchmarking_2025}. Our research instead asks: how do decoder-level uncertainties and hypothesis characteristics manifest differently between resource tiers even before a full word or sentence is output? This sub-token perspective is crucial for understanding the foundational biases and uncertainties that may contribute to downstream performance gaps and for developing targeted interventions.

\section{Data}
To investigate Whisper's sub-token decoding characteristics across different linguistic contexts and resource availability, we curated a diverse set of 20 languages. These languages were categorized into three tiers, High, Medium, and Low resource for better visualization, based on their representation in Whisper's own training data composition \citep{radford_robust_2023}. It should be noted that the definition does not correspond to actual resource levels in the real world beyond Whisper's training dataset. The selected languages are detailed in Table~\ref{tab:whisper_training_hours}. English, while being the most represented language in Whisper's training data, was intentionally excluded from our analysis. Its training data volume is disproportionately larger (over 430,000 hours) compared to the other high resource languages analyzed (e.g., German with approximately 13,000 hours), which would skew the comparative analysis across resource tiers and make it less informative for studying graduated cross-linguistic differences.

To maintain consistency in sub-token analysis and simplify cross-linguistic comparisons at the sub-token level, our analysis primarily focuses on a subset the languages supported by Whisper and available on Common Voice. We restricted our scope to languages written in the Latin script. Pilot analyses of non-Latin scripts (e.g., Chinese, Japanese, Tibetan, Amharic) revealed systematic transcription failures such as outputs rendered in Latin characters rather than the intended script, which would have confounded our cross-linguistic comparisons and PCA analysis. Additionally, we made an explicit decision to exclude languages for which our initial baseline ASR performance using Whisper-large-v2 was exceptionally poor (specifically, WER higher than 60\%). Preliminary qualitative analyses on languages such as Uzbek, Swahili and higher resource Danish revealed that the model frequently misrecognized the target language entirely or produced outputs with non-canonical orthography, making a meaningful analysis impractical.

For each of the selected languages, we used approximately 10 minutes of speech data randomly sampled from the validated subset in the Common Voice 17.0 dataset \citep{ardila_common_2020}. To assess whether 10 minutes adequately represents the sub-token space, we computed a token coverage curve on held-out Estonian: cumulative unique decoder sub-tokens observed as duration increases (1×10min … 6×10min = 60min). The first 10 minutes cover \(\sim\)73\% of the unique sub-tokens that appear by 60 minutes; 30 minutes reach \(\sim\)89\%, and 40 minutes \(\sim\)94\%, after which gains are \(\le\)6\% per additional 10 minutes. This motivates our use of 10 minutes as a practical representative sample (frequent tokens dominate decoder behaviour), while we revisit duration choice in Section \ref{sec:limitations}.

\begin{table}[t]
  \centering
  \small
  \begin{tabular}{lrr}
    \toprule
    \textbf{Resource Tier} & \textbf{Language} & \textbf{Training Hours} \\
    \midrule
    High & German & 13,344 \\
     & Spanish & 11,100 \\
     & French & 9,752 \\
     & Portuguese & 8,573 \\
     & Turkish & 4,333 \\
    \midrule
    Medium & Italian & 2,585 \\
     & Swedish & 2,119 \\
     & Dutch & 2,077 \\
     & Catalan & 1,883 \\
     & Finnish & 1,066 \\
     & Indonesian & 1,014 \\
     % & Vietnamese & 691 \\
     & Hungarian & 379 \\
     & Romanian & 356 \\
     & Norwegian & 266 \\
     % & Czech & 192 \\
    \midrule
    Low & Welsh & 73 \\
     & Lithuanian & 67 \\
     & Latvian & 65 \\
     & Azerbaijani & 47 \\
     & Estonian & 41 \\
     & Basque & 21 \\
    \bottomrule
  \end{tabular}
  \caption{Whisper training hours by language, categorized by resource level.}
  \label{tab:whisper_training_hours}
\end{table}

\section{Methodology}
\subsection{Sub-token Extraction}
Our methodology involves two key stages: generating hypothesis transcriptions and capturing the decoder's state at each generation step. For each audio utterance, we first obtain a transcription using Whisper-large-v2 with beam search (beam size=5, temperature=0.2). We provide the correct language ID token in the initial prompt to ensure transcription in the target language. This process yields the hypothesis sequence of sub-tokens $C = (c_1, c_2, \ldots, c_{N_H})$.

Once this hypothesis is generated, we re-trace its generation path step-by-step. For each position $s$ in sequence $C$, we:
\begin{enumerate}
    \item Provide the decoder with the audio features (encoder output) and the prefix of already chosen sub-tokens $(c_1, c_2, \ldots, c_{s-1})$
    \item Extract the decoder's full probability distribution over all possible sub-tokens for step $s$
    \item Record the top-$K_{cand}=50$ sub-token candidates $T_{s,K_{cand}} = (t_{s,1}, \ldots, t_{s,K_{cand}})$ along with their respective log-probabilities
\end{enumerate}

This re-tracing procedure captures the decoder's internal state—its ranked candidates and their probabilities—at each decision point in the transcription process. These snapshots form the basis for calculating all our analytical metrics and provide insight into the model's decision-making across different languages.

\subsection{Metrics}
We compute several metrics by analyzing the beam search path of hypotheses generated by Whisper. For each utterance, we denote $C = (c_1, c_2, \ldots, c_{N_H})$ as the sequence of $N_H$ sub-tokens in the hypothesis. At each step $s$ in the generation process:
\begin{itemize}
    \item $c_s$ is the sub-token selected by beam search
    \item $T_{s,K_{cand}} = (t_{s,1}, t_{s,2}, \ldots, t_{s,K_{cand}})$ represents the top-$K_{cand}$ (50 in our experiments) sub-token candidates predicted by the decoder, with corresponding probabilities $(p_{s,1}, p_{s,2}, \ldots, p_{s,K_{cand}})$
\end{itemize}

For metrics requiring comparison against ground truth, we use the reference transcription $G = (g'_1, g'_2, \ldots, g'_{N_R})$, tokenized with Whisper's tokenizer. Metrics are aggregated over all relevant items for a given language $L$, with $N_{H_L}$ denoting the total sub-tokens generated across all hypotheses and $N_{R_L}$ the total sub-tokens in all reference transcriptions for language $L$.

\subsubsection{Average Rank of Correct Sub-token}
This metric assesses how highly the ground truth sub-tokens rank among the decoder's predictions. We use Levenshtein alignment to map each reference sub-token $g'_k$ to a hypothesis sub-token $c_s$ (or identify it as a deletion). For each reference sub-token $g'_k$:
\begin{itemize}
    \item If $g'_k$ is aligned with hypothesis sub-token $c_s$: The rank of $g'_k$ is its 1-indexed position in $T_{s,K_{cand}}$. If $g'_k$ is not found within the top-$K_{cand}$ list, its rank is assigned a penalty value of $K_{cand} + 1$.
    \item If $g'_k$ is deleted: Its rank is also assigned the penalty value $K_{cand} + 1$.
\end{itemize}

The average rank for a language $L$ is the mean of these individual ranks across all reference sub-tokens:
$$ \overline{\text{Rank}}(g')_L = \frac{1}{N_{R_L}} \sum_{k=1}^{N_{R_L}} R(g'_k) $$
A lower average rank indicates that the correct sub-tokens are more frequently found among the model's top predictions.

To illustrate this process, Table~\ref{tab:levenshtein_alignment} shows the alignment between ground truth and model-generated sub-tokens for a Turkish phrase. The table demonstrates possible alignment operations: equal (where the model correctly identified the token), replace (where the model chose a different token), and delete (where a ground truth token has no corresponding model token). 

\begin{table}[t]
\centering
\small
\begin{tabular}{ccccc}
\toprule
\textbf{Position} & \textbf{GT} & \textbf{Output} & \textbf{Operation} & \textbf{Rank} \\
\midrule
0 & $\langle$|BOS|$\rangle$ & $\langle$|BOS|$\rangle$ & equal & 1 \\
1 & Sil & S & replace & 1 \\
2 & ah & el & replace & 5 \\
3 & lar & am & replace & 7 \\
4 & … & lar & replace & 4 \\
5 & $\langle$|EOS|$\rangle$ & (deleted) & delete & $K+1$ \\
6 &  & A & insert & -- \\
7 &  & ley & insert & -- \\
8 &  & kum & insert & -- \\
9 &  & . & insert & -- \\
10 &  & $\langle$|EOS|$\rangle$ & equal & 2 \\
\bottomrule
\end{tabular}
\caption{Alignment between ground truth (GT) sub-tokens (``Silahlar…''/``Weapons…'') and model output sub-tokens (``Selamun Aleykum.''/``Peace be upon you.'').}
\label{tab:levenshtein_alignment}
\end{table}

For this example, the average rank is calculated as: $\overline{\text{Rank}} = \frac{1 + 1 + 5 + 7 + 4 + (K+1) + 2}{7} = \frac{20 + (K+1)}{7}$ = 10.14 with $K=50$. 

\subsubsection{Confidence}
Confidence measures the average probability assigned to the sub-token $c_s$ that was ultimately chosen at each step along the beam search path:
$$ \overline{\text{Conf}}_L = \frac{1}{N_{H_L}} \sum_{s=1}^{N_{H_L}} p(c_s | \text{utterance}, c_{<s}) $$
where $p(c_s | \text{utterance}, c_{<s})$ is the probability assigned to $c_s$ given the utterance audio input and previous tokens.

\subsubsection{Entropy}
Token-level entropy quantifies the uncertainty in the model's predictions, calculated over the top-$K_H=50$ predicted sub-tokens. After normalizing the probabilities of these candidates, the entropy (in bits) at step $s$ is:
$$ H_s = - \sum_{i=1}^{K_H} p'_{s,i} \log_2 p'_{s,i} $$
where $p'_{s,i}$ are the renormalized probabilities over the top $K_H$ candidates. The average entropy $\bar{H}_L$ for a language is reported as the mean across all decoding steps.

\subsubsection{Alternate-candidate Diversity}
This metric evaluates the variety within the set of predicted candidates using the Type-Token Ratio (TTR). Specifically, we calculate the TTR of the non-top-1 candidates within the top-$K_D=50$ predictions. For each language, we collect all sub-tokens that appear as candidates ranked from 2 to $K_D$ across all decoding steps. The diversity is then computed as:
$$ \text{Diversity}_L = \frac{|\text{unique non-top-1 tokens in L}|}{|\text{total non-top-1 tokens in L}|} $$

This approach explores the richness of the hypothesis space beyond the model's single best guess. Our underlying assumption is that lower-resourced languages might exhibit less diversity in these alternative candidates, potentially reflecting a more constrained or less nuanced hypothesis space learned by the model due to limited training data. A higher TTR indicates a broader range of unique alternatives being considered.

\subsubsection{Sub-token Use Visualization}
To visualise cross-lingual patterns in sub-token usage, we build a frequency vector for each language from the sub-tokens that appear among the top-$K_{\text{PCA}} = 10$ candidates at every decoding step. These frequency vectors represent the distribution of sub-token IDs from Whisper's multilingual vocabulary of 51,865 tokens \citep{radford_robust_2023}. Using a wider window (e.g.\ $K_{\text{PCA}} = 50$) would fold in many low-frequency alternates and blur fine-grained distinctions, so we fix $K$ at 10 to keep language differences salient.  

After standardizing these vectors, we apply two dimensionality reduction techniques: Principal Component Analysis (PCA) and t-distributed Stochastic Neighbor Embedding (t-SNE). PCA projects the data onto the first two principal components to reveal linear relationships and global structure, while t-SNE emphasizes local neighborhoods and non-linear clustering patterns. The two approaches allow us to examine both broad cross-linguistic trends and fine-grained language-specific patterns in sub-token usage. For t-SNE, We used scikit-learn t-SNE with perplexity set to 20 (clamped automatically to $(n-2)/3$ if the number of languages was small).

\section{Results}
\begin{figure}[t]
  \centering
  \includegraphics[width=\linewidth]{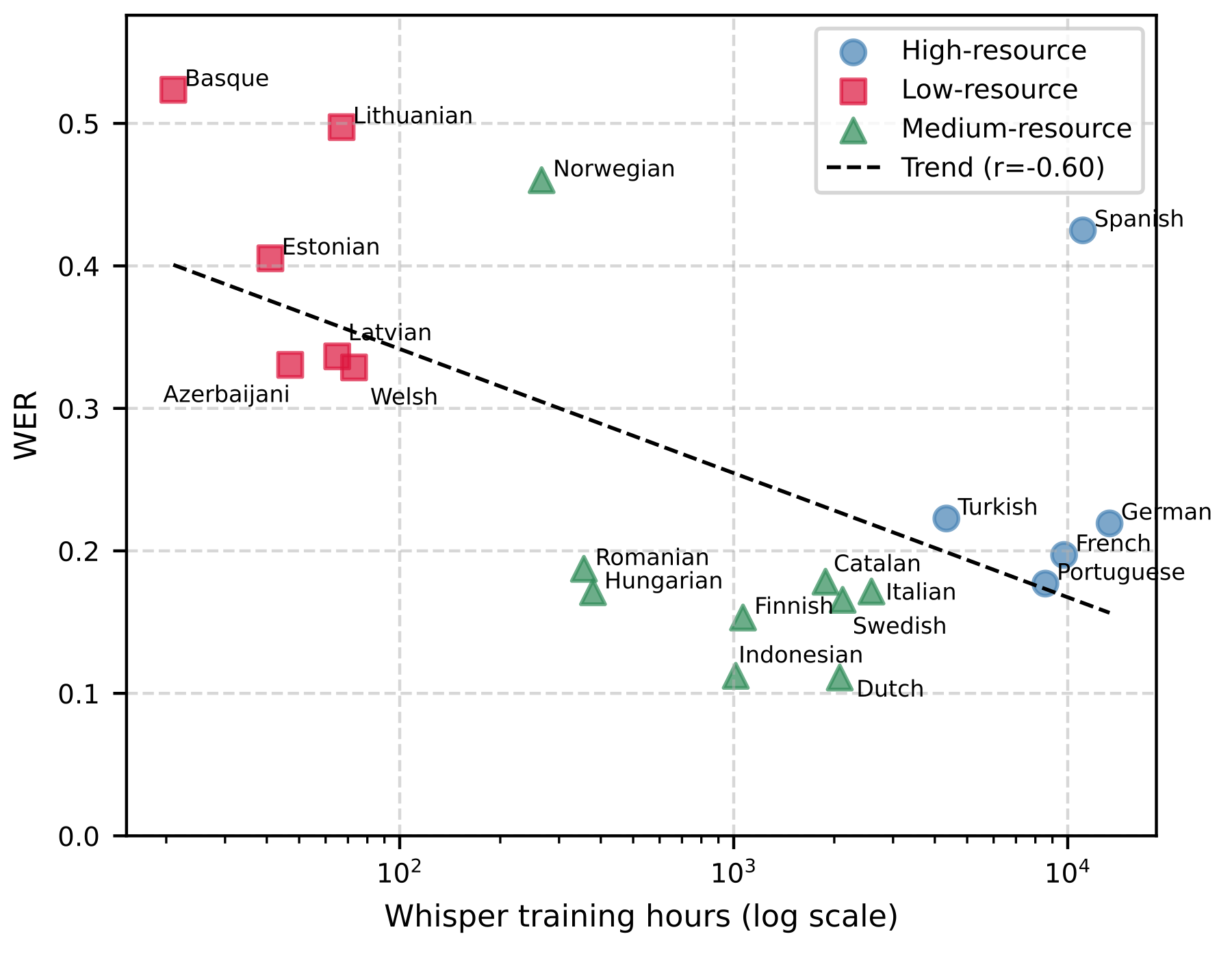}
  \caption{WER of Whisper on the Common Voice dataset versus training hours. Higher resource languages tend to have lower WER ($p$-value < 0.001). }
  \label{fig:wer}
\end{figure}

Our analysis of Whisper's sub-token decoder reveals systematic variations in its behavior that correlate strongly with language resource levels, quantified by training hours. As a baseline, Figure \ref{fig:wer} reports the Word Error Rate (WER) of the languages studied.

Consistent with previous findings we observe that WER is generally lower for languages with more training hours \citep{radford_robust_2023}. Our subsequent sub-token level analyses aim to delve deeper into the internal decoding characteristics that might contribute to these performance differences.

\subsection{Rank of Correct Sub-token}
\begin{figure}[t!] 
  \centering
  \includegraphics[width=\linewidth]{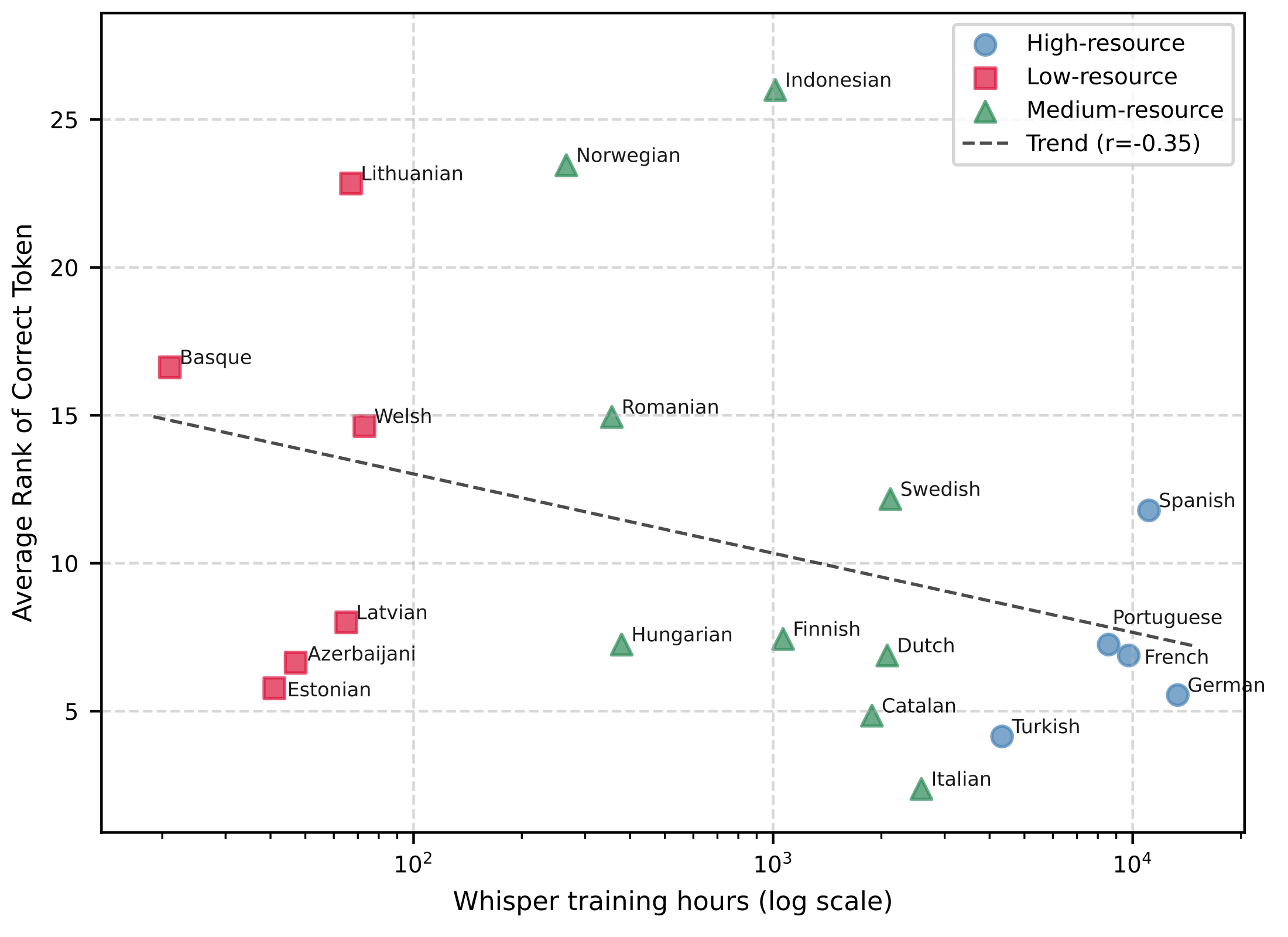} 
  \caption{Average rank of the correct sub-token versus Whisper training hours. Higher resource languages tend to have correct tokens ranked higher (closer to 1, $p$-value < 0.001).}
  \label{fig:rank_chosen}
\end{figure}

The average rank of the correct sub-token correlates with language resource levels, as shown in Figure~\ref{fig:rank_chosen}. Higher resource languages have their correct sub-tokens ranked higher. This indicates that for higher resource languages, the beam search predominantly follows the locally highest-probability path. Despite the general trend, we do observe some deviations in individual languages. 

\subsection{Decoder Confidence and Predictive Entropy}
\begin{figure}[t]
  \centering
  \includegraphics[width=\linewidth]{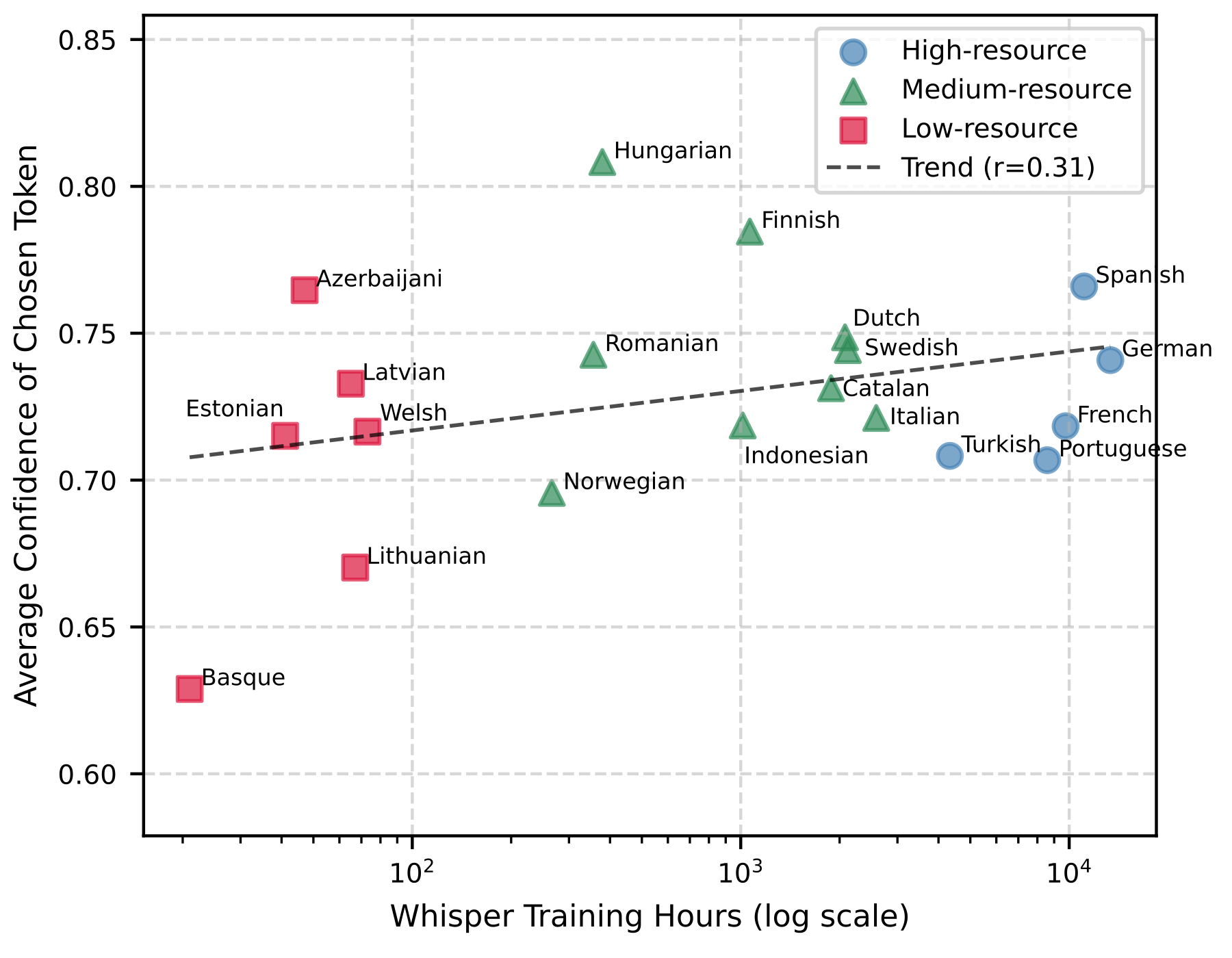} 
  \caption{Average model confidence (probability of chosen sub-token) versus Whisper training hours. Higher-resource languages generally exhibit higher confidence ($p$-value < 0.001).}
  \label{fig:confidence}
\end{figure}

Decoder confidence, measured as the average probability of the chosen sub-token, shows a strong positive correlation with language training hours, as shown in Figure~\ref{fig:confidence}. High-resource languages tend to exhibit higher average confidence values. 

\begin{figure}[t]
  \centering
  \includegraphics[width=0.9\linewidth]{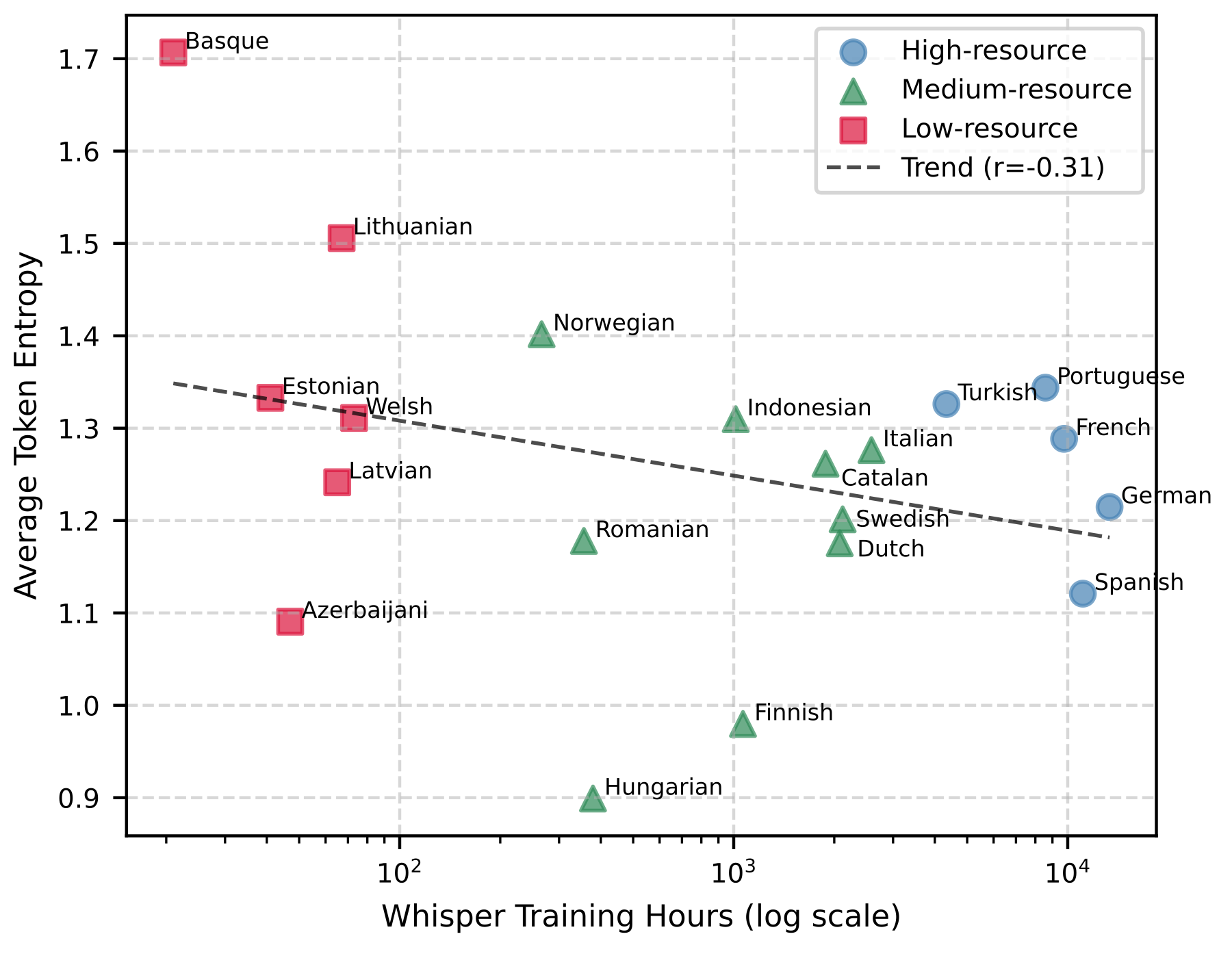} 
  \caption{Average predictive token entropy of the top-50 candidates versus Whisper training hours. Higher resource languages generally exhibit lower entropy ($p$-value < 0.001).}
  \label{fig:entropy}
\end{figure}

Conversely, predictive entropy, calculated over the top-5 predicted sub-tokens, demonstrates a negative correlation with training hours, as shown in Figure~\ref{fig:entropy}. High-resource languages show lower average entropy, indicating more peaked and certain predictive distributions. This inverse relationship between confidence and entropy is expected: when the model is more confident in its chosen token, its distribution over alternatives is sharper (less entropic). 

\subsection{Alternate‑Candidate Diversity}
\begin{figure}[t!]
  \centering
  \includegraphics[width=\linewidth]{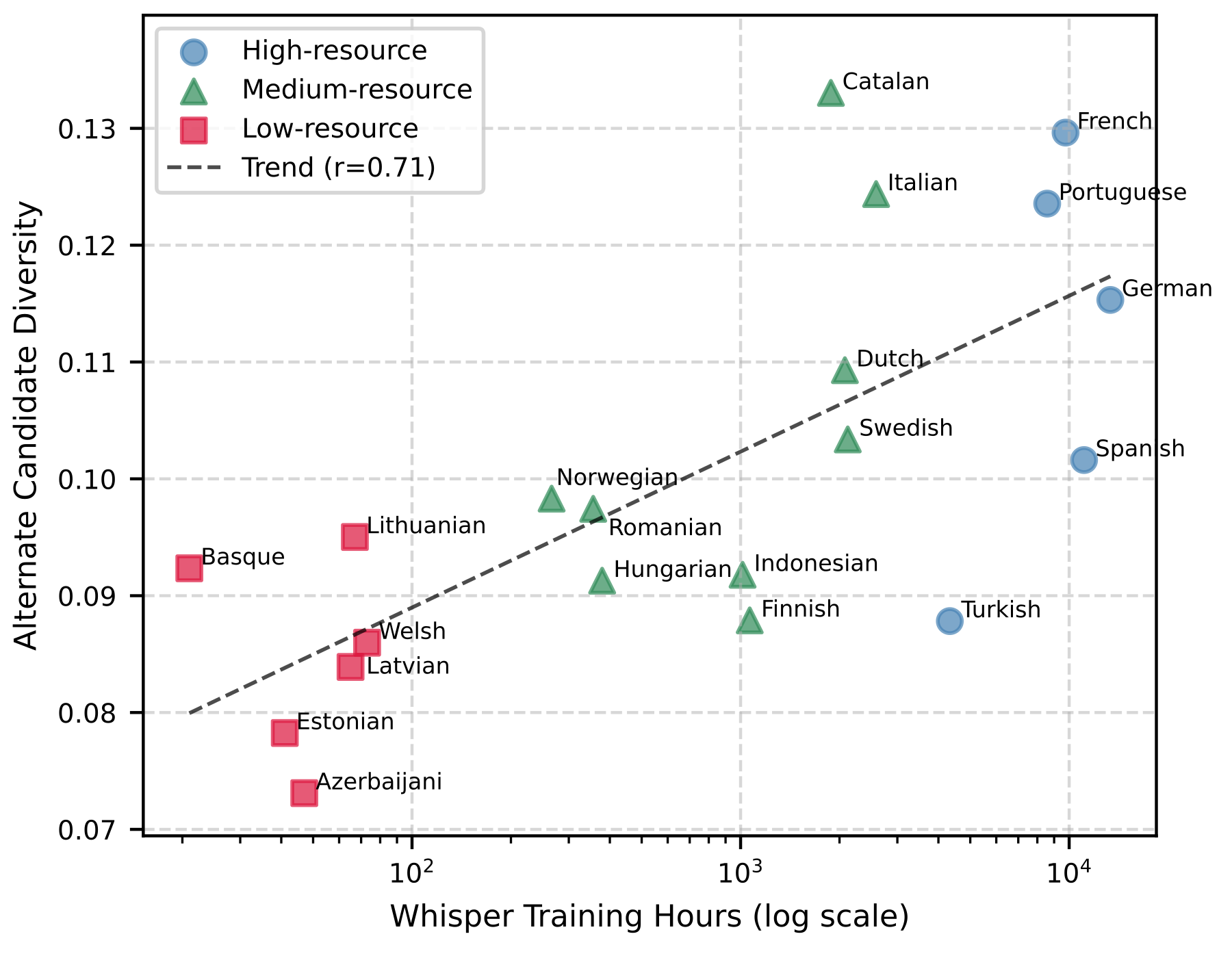}
  \caption{Alternate-candidate diversity (TTR of non-top-1 candidates in top-50 candidates) versus Whisper training hours. Higher resource languages tend to have higher alternate-candidate diversity ($p$-value < 0.001).}
  \label{fig:diversity}
\end{figure}

Alternate-candidate diversity, measured as the TTR of non-top-1 candidates within the top-5 predictions, exhibits a generally positive correlation with language resource levels, as shown in Figure~\ref{fig:diversity}. Higher resource languages tend to populate the upper range of diversity scores. This suggests that for higher resource languages, the model often considers a richer set of unique alternatives beyond its top choice. The overall trend indicates that increased training data may lead to a more varied set of hypotheses being considered.

\subsection{PCA and t-SNE Clustering of Sub‑token Usage}
\begin{figure}[t!]
  \centering
  \includegraphics[width=\linewidth]{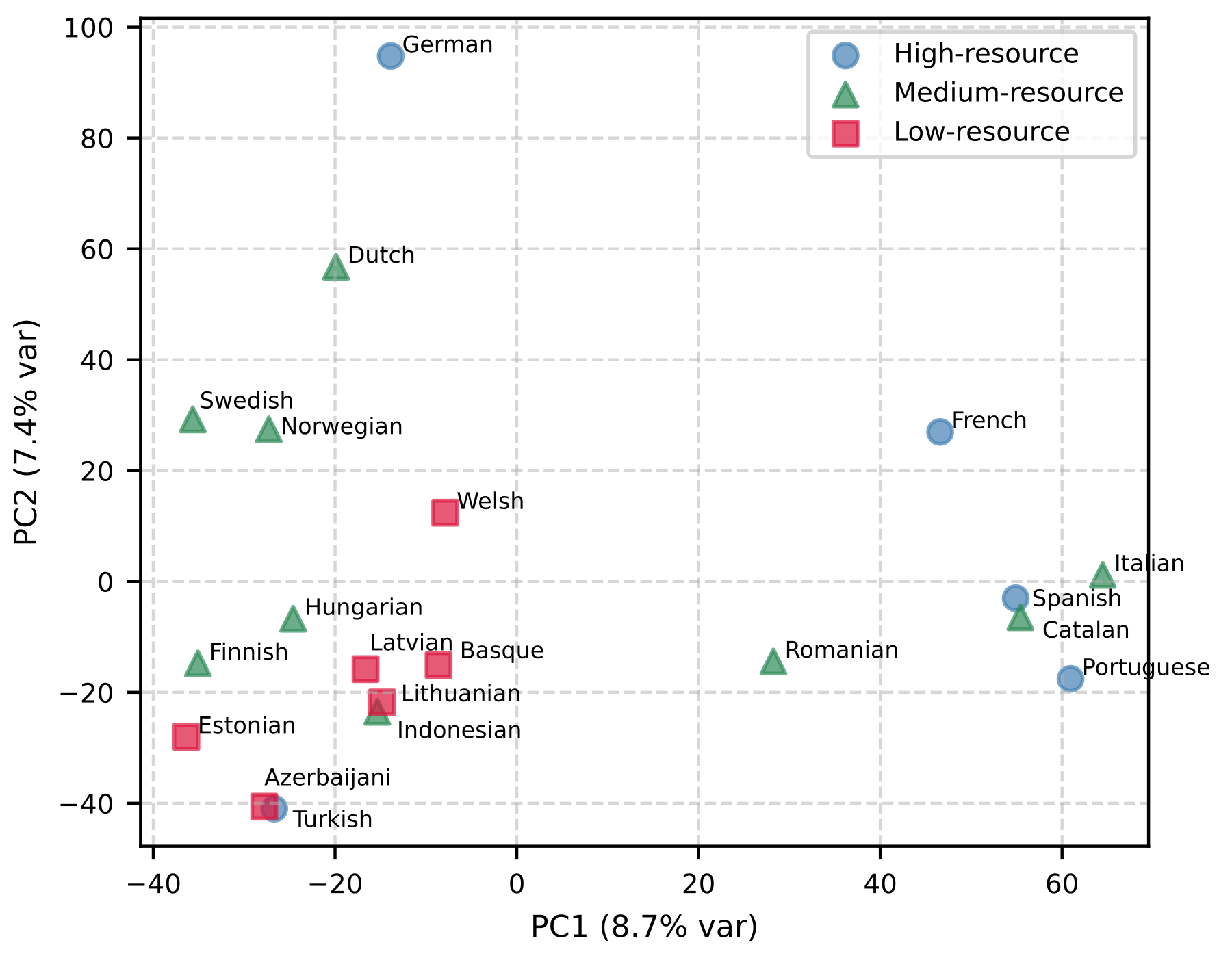} 
  \caption{PCA of sub‑token usage frequency vectors (top-10 candidates).}
  \label{fig:pca}
\end{figure}

The PCA visualization of language-specific sub-token frequency vectors (from top-10 candidates) reveals distinct clustering patterns that reflect both typological relationships and resource levels, as presented in Figure~\ref{fig:pca}. A prominent observation is the separation of language families. Romance languages (Spanish, French, Portuguese, Italian, Catalan, and Romanian) form a relatively cohesive cluster. Similarly, Turkic languages (Turkish and Azerbaijani), North Germanic languages (Swedish and Norwegian), Uralic languages (Finnish and Estonian, and to a lesser extent Hungarian), tend to stay clustered closer together and distinctly separated from other groups along the principal components. The clustering is also interesting as it groups languages of varying resource levels together, suggesting that shared linguistic structures (e.g., common morpho-phonological features captured in common sub-tokens) could influence sub-token usage patterns.

Notably, typologically unrelated lower resource languages, such as Welsh, Lithuanian, Latvian, and Basque, cluster together despite their significant linguistic differences. This unexpected grouping suggests these languages are handled similarly by Whisper's decoder not because of linguistic similarity, but due to their shared status as low-resource languages in the training data. Unlike high-resource languages that form distinct family-based clusters, these unrelated low-resource languages appear to share common sub-tokenization characteristics that transcend actual linguistic relationships, indicating deficiencies in the model's sub-token representations where it may be falling back to more generic decoding patterns due to insufficient exposure during training.

\begin{figure}[t!]
  \centering
  \includegraphics[width=\linewidth]{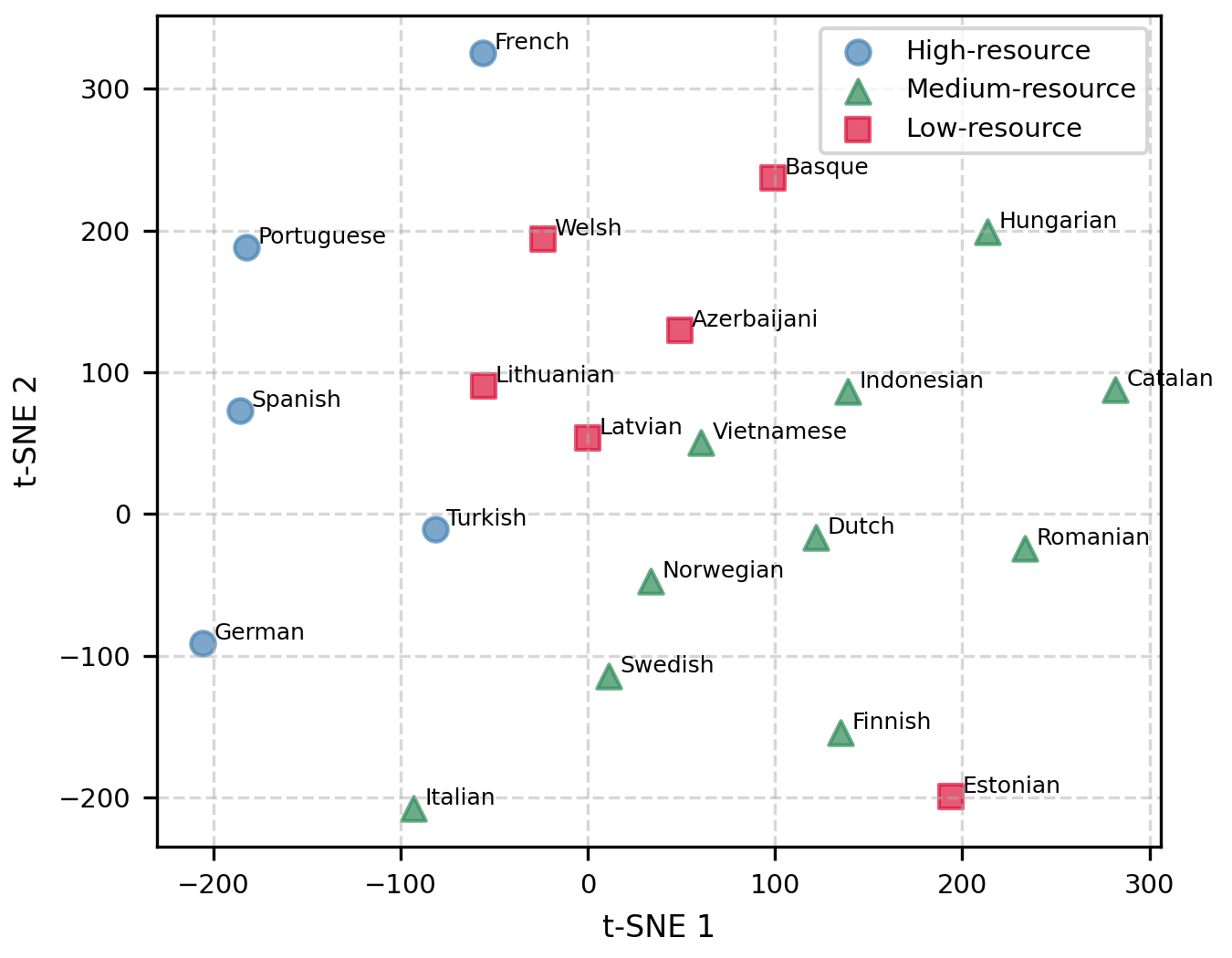} 
  \caption{t-SNE of sub‑token usage frequency vectors (top-10 candidates).}
  \label{fig:tsne}
\end{figure}

The t-SNE analysis (Figure~\ref{fig:tsne}) provides additional insights into local clustering patterns while confirming several key observations from the PCA. Several neighborhoods remain stable across both dimensionality reduction techniques, particularly Lithuanian–Latvian, Finnish–Estonian, and Norwegian–Swedish pairs, suggesting robust family- or script-level commonalities in sub-token usage. However, t-SNE reveals that family structure is weaker when emphasizing local neighborhoods: Romance languages do not form a single tight cluster as they do in PCA, with Spanish/Portuguese/French appearing more separated. This suggests that while broad cross-lingual trends are largely linear and captured effectively by PCA, fine-grained, language-specific token frequencies can create local distinctions that pull related languages apart in the t-SNE embedding.

The resource-linked patterns observed in PCA are reinforced in t-SNE, where medium-resource languages occupy several tight neighborhoods, while unrelated low-resource languages (Welsh, Lithuanian, Latvian, Azerbaijani) continue to cluster together. This consistency across both methods strengthens the interpretation of a "generic decoding" regime under data scarcity. Notably, Estonian and Basque emerge as outliers in both visualizations, suggesting these languages exhibit particularly idiosyncratic sub-tokenization patterns that warrant further investigation.

\section{Discussion}

Our sub-token probing of Whisper's decoder uncovers systematic variations in its behavior. These variations not only correlate strongly with language resource levels but are also significantly shaped by linguistic typology, offering a more nuanced understanding of the model's internal mechanisms beyond aggregate error rates.

\subsection{Resource-Driven Variations in Sub-token Decoding}
The analysis consistently reveals that higher resource languages benefit from more favorable decoding characteristics. Specifically, the model exhibits higher average confidence in its chosen sub-tokens for higher resource languages, and their predictive distributions are marked by lower entropy, indicating greater certainty in its predictions. Furthermore, the correct sub-token is more frequently highly ranked among the candidates considered during beam search for these languages. 

These advantages can be attributed to the extensive exposure to diverse linguistic phenomena afforded by larger training datasets \citep{radford_robust_2023}. Such exposure likely enables the formation of more robust and well-calibrated sub-token representations. The generally higher alternate-candidate diversity observed for higher resource languages also suggests that increased training data allows the model to consider a richer and more varied set of plausible alternatives during the decoding process, potentially contributing to improved overall transcription accuracy.

\subsection{The Influence of Linguistic Typology on Sub-token Usage}
Beyond the volume of training data, both the PCA and t-SNE analyses of sub-token usage (Figures~\ref{fig:pca} and~\ref{fig:tsne}) illustrate the strong influence of linguistic structure on the model's internal representations. Typological clusters observed across both methods—such as Lithuanian–Latvian, Finnish–Estonian, and Norwegian–Swedish—underscore that shared morpho-syntactic or phonological features drive stable sub-token patterns.

PCA and t-SNE offer complementary perspectives. PCA captures broad linear trends, with Romance languages forming a cohesive family cluster, while t-SNE emphasizes local neighborhoods where fine-grained token frequency differences separate closely related languages (e.g., Spanish, Portuguese, and French). Typological relationships thus operate at multiple scales: family-level similarities captured by linear methods, and language-specific idiosyncrasies revealed by non-linear analysis.

Clustering patterns also group higher- and lower-resource languages by typological relatedness. Inherent properties such as agglutinative morphology can shape representational space, partly mitigating data scarcity disadvantages when low-resource languages belong to related families. Typological relatedness therefore conditions sub-token representations and may support cross-lingual transfer.

At the same time, unrelated low-resource languages often cluster together despite their differences, suggesting Whisper's decoder treats them similarly due to shared scarcity rather than linguistic similarity. The persistence of this effect in both PCA and t-SNE indicates that data scarcity shapes representations in ways independent of the analysis method.

Finally, Estonian and Basque consistently emerge as outliers, reflecting idiosyncratic sub-tokenization patterns that may require targeted investigation and specialized handling in multilingual ASR systems.

\subsection{Sub-token Prediction Accuracy versus Global Performance}
An intriguing aspect of our findings is the observed incongruity between local sub-token prediction and global Word Error Rate (WER) for particular languages. For instance, languages like Estonian and Azerbaijani, despite demonstrating remarkably low average ranks for their gold sub-tokens (signifying that the correct orthographic units are included as high-probability candidates by the acoustic model at a local, per-step level) do not invariably achieve the lowest overall WERs in our analyzed set.

This phenomenon highlights the inherent complexities of the end-to-end ASR decoding pipeline. While the model may possess robust local "knowledge" of correct sub-units, the ultimate transcription quality is a cumulative function of global sequence optimization during beam search, the effective mitigation of cascading errors arising from any single misstep, and the nuanced handling of intricate linguistic features that span multiple tokens. Consequently, for such languages, interventions aimed solely at further refining local sub-token prediction accuracy might yield diminishing returns on WER improvement. Strategies that enhance the model's capacity for global context modeling or its specific handling of overarching linguistic complexities may prove more fruitful.

\subsection{Practical Implications}
The above analysis offer several avenues for targeted interventions. The identification of language-specific weaknesses, such as a consistent tendency for correct sub-tokens to be ranked lower or for predictive distributions to exhibit high entropy in low resource languages, can directly inform the design of language-specific adapters or more sophisticated parameter-efficient fine-tuning strategies \citep{song_lora-whisper_2024, pfeiffer_unks_2021}. Furthermore, decoder-internal metrics, including dynamic measures of confidence and entropy, could potentially serve as valuable signals for adaptive adjustments to decoding algorithms. Finally, the sub-token usage patterns unveiled by PCA and the diversity metric may illuminate critical deficiencies in current vocabulary coverage and suggest novel data augmentation techniques. 

\section{Conclusion}
This study introduces a fine-grained sub-token probing framework for Whisper's decoder, revealing systematic disparities in how the model processes languages across resource levels. Higher resource languages consistently benefit from more favorable decoding characteristics: correct tokens more frequently top-ranked, higher confidence, lower entropy, and more diverse alternative candidates. Our complementary PCA and t-SNE analyses of sub-token usage further demonstrate that linguistic typology significantly influences these representations, with related languages clustering together regardless of resource tier, while unrelated low-resource languages unexpectedly cluster due to similar sub-tokenization patterns rather than linguistic similarity. These insights, often masked by aggregate metrics like WER, highlight how resource disparities manifest within the decoder's internal mechanisms and point toward targeted interventions such as language-specific adapters, dynamic decoding strategies, and focused data augmentation to promote more equitable multilingual ASR development.

\section*{Limitations}
\label{sec:limitations}
Although we tried to present a comprehensive analysis, our study has several limitations. First, we focused primarily on languages using the Latin script, excluding many writing systems and potentially missing script-specific tokenization effects. We also omitted languages with extremely high error rates (WER > 60\%), where Whisper frequently misidentified the language altogether, making sub-token analysis unreliable. Additionally, our 10-minute samples per language provide only a snapshot that may not capture the full phonological diversity or domain variation present in natural speech. 

Our methodology for extracting and evaluating sub-token performance also relies on assumptions about alignment between hypothesis and reference transcriptions that may not perfectly represent ground truth. While we analyze the chosen beam search path, we do not directly probe Whisper's internal beam search heuristics or alternate paths, which might offer additional insights into model decision-making. Finally, our resource-level categorization is based solely on Whisper's training hours rather than real-world language resources. 

Furthermore, our analysis is inherently dependent on the quality and accuracy of the labels provided within the Whisper training dataset. As noted by \citet{radford_robust_2023} themselves, there can be instances of labeling errors within this large dataset (e.g., some English audio being mislabeled as Welsh). Similar mislabelings for other languages could exist and potentially influence the model's learned representations and, consequently, our observations, particularly for languages where such noisy data might constitute a non-negligible portion of their training subset.

Future work should extend this analysis to more scripts, longer and more diverse audio samples, and explore how sub-token behavior correlates with specific linguistic features across typologically diverse languages.

\section*{Acknowledgements}
Part of this research was funded by the University of Washington (UW) Global Innovation Fund (GIF) Research award for a collaboration between UW and Université Paris Cité\footnote{\url{https://altae.u-pariscite.fr/project/gifra/}} to promote fairness for under-represented languages in multilingual LLMs (Promoting fairness for under-represented languages in multilingual LLMs project (2025-2026). 
% other grants ?

% Entries for the entire Anthology, followed by custom entries
\bibliography{anthology,custom}
\bibliographystyle{acl_natbib}

% \appendix
% \section{Example Appendix}
% \label{sec:appendix}

\end{document}